# A transformer-based spelling error correction framework for Bangla and resource scarce Indic languages

Mehedi Hasan Bijoy[1,†], Nahid Hossain[2,†], Salekul Islam[2,3] and Swakkhar Shatabda[2,3]
[1] Department of Information and Communications Engineering, Aalto University
[2]Department of Computer Science and Engineering, United International University
[3]Center for Artificial Intelligence and Robotics (CAIR), United International University
[†] Joint First authors

**Abstract**

Spelling error correction is the task of identifying and rectifying misspelled words in texts. It is a potential and active research topic in Natural Language Processing because of numerous applications in human language understanding. The phonetically or visually similar yet semantically distinct characters make it an arduous task in any language. Earlier efforts on spelling error correction in Bangla and resource-scarce Indic languages focused on rule-based, statistical, and machine learning-based methods which we found rather inefficient. In particular, machine learning-based approaches, which exhibit superior performance to rule-based and statistical methods, are ineffective as they correct each character regardless of its appropriateness. In this paper, we propose a novel detector-purificator-corrector framework, DPCSpell based on denoising transformers by addressing previous issues. In addition to that, we present a method for large-scale corpus creation from scratch which in turn resolves the resource limitation problem of any left-to-right scripted language. The empirical outcomes demonstrate the effectiveness of our approach, which outperforms previous state-of-the-art methods by attaining an exact match (EM) score of 94.78%, a precision score of 0.9487, a recall score of 0.9478, an f1 score of 0.948, an f0.5 score of 0.9483, and a modified accuracy (MA) score of 95.16% for Bangla spelling error correction. The models and corpus are publicly available at https://tinyurl.com/DPCSpell.

## 1 Introduction

A survey shows that 89.3% of native speakers make spelling errors while writing an essay, whereas it increases to 97.7% for non-native speakers [1]. The goal of spelling error rectification is to automatically detect and correct spelling mistakes in the text. It is an important task in Natural Language Processing (NLP) to improve the performance of numerous downstream activities such as machine translation, text generation and summarization, sentiment analysis, and web search engines, to name a few. However, spelling errors are usually caused by the wrong placement of phonologically and visually identical letters, as well as typing mistakes while writing. Hence, spelling error correction methods necessitate a complete grasp of word similarity in terms of phonetic sounds and visual shapes along with typing patterns, language reasoning, and modeling, which in turn makes it one of the most challenging tasks in NLP.

Many approaches have been proposed for spelling error correction (SEC) which are mostly language-specific. Furthermore, almost all of these approaches rely on a tiny corpus that is not publicly accessible, making reproducibility miserable. Besides, they are restricted to correcting only a few types of errors. However, we emphasize Sanskrit-oriented resource-scarce languages such as Bangla, Hindi, and Telugu SEC in this work because they share a left-to-right typing script [2]. Most of the SEC methods of these languages are either based on heuristic rules [3, 4] or conventional language models [5, 6]. Recently, with the emergence of NLP, some methods have been proposed utilizing the Recurrent Neural Network (RNN) based sequence-to-sequence approach for SEC of resource-limited languages [7]. These methods correct each character of the word regardless of its correctness, which might affect the correct characters and lead to a high type-I error rate. Even if they are able to rectify the word, doing so would require them to unnecessarily correct letters that are already accurate. When only a tiny portion of words are erroneous, which is common in spelling errors, this problem becomes much more severe. However, since none of the existing studies utilize transformer-based methods in relation to the SEC task of Bangla, Hindi, and Telugu languages, we undertake a comprehensive scrutiny to assess and authenticate the untapped capabilities of transformers.

In this article, we propose a novel denoising transformer-based detector-purificator-corrector framework for SEC of Bangla and low-resource Indic languages by addressing the above issues and name it DPCSpell. Unlike previous methods, it corrects the erroneous characters only. The DPCSpell comprises three networks: a detector, a purificatior, and a corrector. Firstly, the detector module takes the erroneous word as input and masks the wrong characters using a transformer-based method. Secondly, a similar transformer-based model is employed to further purify the masks of the detector module as the correction largely relies on the masked characters. Finally, the original erroneous input and masked output from the purificator are fed into the corrector module which synthesizes the correction. The whole approach is illustrated in Figure 2. Furthermore, we propose a method for creating a large-scale parallel corpus that resolves the resource-limitation issue for SEC of any left-to-right scripted language such as Bangla, Hindi, or Telugu. Especially, a large-scale parallel corpus for Bangla SEC is developed using our method and made publicly available. Likewise, the Hindi and Telugu corpora are enhanced following our method. The empirical outcomes elucidate the efficacy of our method in the SEC and the fruitfulness of our corpus creation approach. Additionally, we promote transparency by making all our codes publicly available, fostering reproducible baseline of the task.

The contributions of this article are summarized below:

- We propose a novel detector-purificator-corrector framework named DPCSpell, which is based on denoising transformers, for the SEC of Bangla and resource-scarce Indic languages such as Hindi and Telugu.
- We compare our method with state-of-the-art methods in different languages. It has become the new state-of-the-art method for Bangla SEC.
- A comprehensive comparison among rule-based, RNN-based, convolution-based, and transformer-based methods is performed for the SEC task.
- We introduce a method for developing a large-scale parallel corpus from scratch that overcomes

the resource scarcity issue of left-to-right scripted languages. A large-scale parallel corpus for Bangla SEC is developed using our method and made publicly available, making Bangla no longer a low-resource language for the SEC task.

The rest of the article is organized as follows: Section 2 contains a thorough literature review of Bangla SEC along with Hindi and Telugu. Following that, the explanation of Bangla spelling error types and the BanglaSEC corpus creation procedure are provided in Section 3. In section 4, we explain the methodology of our proposed DPCSpell. Next, we discuss the empirical outcomes, and compare quantitative and qualitative results in Section 5. Moreover, we criticize our method in subsection 5.7. Finally, Section 6 concludes our work with future scope.

## 2 Literature Review

A substantial amount of study has been conducted on Bengali spelling detection and correction. The Bengali spell checker is yet to depict accurate performance like spell checkers in western languages, i.e. English. It is an active research topic in Bangla Natural Language Processing (BNLP) because of the diversity of its applications in text generation, text summarization, web search engines, and sentiment analysis, to name a few. Bengali spell-checking methods can broadly be classified into three categories including rule-based, statistical, and deep-learning-based approach.

### 2.1 Rule-Based Methods

Early efforts in Bengali spell checking mostly focus on generating and employing heuristic rules based on morphology, stemming, parts-of-speech, and so on to detect different types of errors. In these approaches, error detection and correction take place in two distinct phases. Most of these tactics utilize a dictionary lookup table to detect the errors [3, 4, 8, 9, 10, 11, 12, 13], except for a few approaches where a string matching algorithm is used [14]. A variety of algorithms including minimum edit distance [3, 4, 8, 12], Levenshtein's edit distance [13], linear search [9], phonetic [11], soundex and metaphone [14], and Nerving's correct spelling suggestion algorithm [10] are employed for correction generation. The majority of these methods are only effective in rectifying trivial mistakes.

A spell checker for transliterated Bangla words has been proposed in [9] using a dictionary lookup table, an amalgamation of linear search and Damerau-Levenshtein minimum edit distance, and linear search for error detection, correction, and conversion respectively. In 2020, Hasan et al. [10] proposed a spell checker by amending peter Nerving's correct spelling suggestion algorithm. A phonetic encoding technique for Bangla considering context-sensitive rules is designed by utilizing edit distance, soundex, and metaphone algorithms [11]. Another Bengali spell checker [12] employs a clustering-based approach that diminishes both search space and time complexity. Saha et al. [14] brings forward a method by incorporating edit distance, soundex, metaphone, and string matching algorithms to identify the erroneous Bangla words and deliver the optimal suggestion. Recently, Ahamed et al. [4] proposes a strategy that leverages Norvig's algorithm to detect errors in Bengali words and Jaro-Winkler distance to generate suggestions and corrections. Hossain et at. [13] tackles the same problem by utilizing edit distance and double methaphone algorithms based on distributed lexicons and numerical suffixes.

The rule-based methods are susceptible to handling facile mistakes including typographical [3, 4, 8, 13, 14], phonetic [8, 12, 13], and cognitive [13, 14] errors with ease. Since they largely rely on linguistic knowledge, the complexity of processing intricate errors rises significantly as it requires a lot of time and effort to manually construct apt rules. After all, they are bound to a few specific rules for rectifying errors. Consequently, they fail to generalize to new test cases.

## 2.2 Statistical Methods

The statistical approach has long been prevalent in Bengali spell-checking studies because of its impressive performance. Moreover, it does not share the drawbacks of rule-based methods as it avoids sole reliance on linguistic knowledge. The spell-checking is carried out based on different characteristics of words through employing word count, frequency, n-gram language model, finite state automata, and so fourth. Similar to rule-based methods, error identification and rectification happen here in two separate phases. However, statistical approaches can further be classified into statistics-based [5, 6, 15, 16, 17, 18, 19] and machine-learning-based [20, 21, 22] methods. These spell-checkers are proficient in terms of non-word errors but abortive in handling real-word errors.

A hybrid method by incorporating edit distance with the N-gram language model has been proposed for detecting and correcting word-level errors [6, 15]. Furthermore, Mittra et al. [15] utilizes the probabilities from the N-gram model to detect sentence-level errors. Gupta et al. [5] and Khan et al. [19] use a dictionary and the N-gram model for error identification and rectification, respectively. A framework has been developed by [16] employing the N-gram language model to create clusters of words. An amalgamation of bi-gram and tri-gram has been examined to detect and correct homophone and real-word errors in [17]. Another spell checker [18] uses a corpus and finite state automata to detect errors and generate relevant suggestions. Urmi et al. [20] presents an unsupervised method to generate a rich Bengali root word dictionary which will essentially aid in the spell-checking task. A Bengali morphological parser has been demonstrated in [21] by exploiting the stemming cluster approach. Sharif et al. [22] employs logistic regression to classify given Bangla texts into suspicious and non-suspicious classes.

The performance of these methods largely relies on data preprocessing and feature engineering which require domain knowledge. Also, they ignore context by not taking word analogies into account while constructing numerical representations. Certainly, the spell checker would function more effectively if it could determine whether the suggestion is appropriate for the context or not.

## 2.3 Deep-Learning-Based Methods

Although rule-based and statistical approaches perform well, the advent of deep learning has the potential to improve performance even further. They rectify errors by considering context which in turn makes these methods meaningful. These approaches are especially useful for correcting real-word errors where the context of the word in relation to the sentence is required.

A hybrid approach has been proposed in [23] by integrating a bi-gram language model with Long Short-Term Memory (LSTM) network to identify and rectify Bangla real-word errors. Sarker et al. [24] mimics a similar methodology for Bangla word completion and sequence prediction. A Gated

Recurrent Unit (GRU) based Recurrent Neural Network (RNN) on N-gram dataset is employed by [25] to anticipate the next word of a given sequence. Islam et al. [26] presents a sequence-to-sequence method for Bangla sentence correction and auto-completion that utilizes LSTM cells in both the encoder and decoder networks. Two Convolutional Neural Network (CNN) based sub-models have recently been proposed by [2] to handle certain properties of Bengali and Hindi words including high inflection, flexible word order, morphological richness, and phonetical spelling errors. Another study [7] proposes a sequence-to-sequence approach for Hindi and Telegu spell checking that uses an attention-based character-level LSTM network in both the encoder and decoder. Singh et al. [27] mimics an analogous approach for developing a Hindi spell checker based on an encoder-decoder network where LSTM and CBOW word embedding are utilized.

While there have been a few attempts to develop spell checkers for resource-scarce Indic languages like Hindi and Telegu, which are Sanskrit originated and share a similar structure to Bangla, by employing Neural Machine Translation (NMT), to the best of our knowledge, no such work has strived yet in relation to the Bengali spell checker. Therefore, in this article, we employ a transformer-based sequence-to-sequence network for the first time in relation to the Bengali spell checker to ensure its performance.

## 2.4 Spell Checkers in Resource Scarce Languages

A variety of spell-checking approaches have been proposed for different low-resource languages including Bangla [2], Hindi [2, 7, 27, 28], Telugu [7], Punjabi [29], Gujarati [30], Azerbaijani [31], Malayalam [32], Urdu [33], Hungarian [34] and Sinhala [32, 35]. An innovative method for automating the correction of spelling errors in Hungarian clinical records is introduced in [34], utilizing a word-based algorithm and a Statistical Machine Translation (SMT) decoder, even in the absence of an orthographically correct proofread corpus from this domain.. A few recent approaches such as [28], [30], and [36] propose a rule-based method for Hindi, Gujarati, and Sinhala spelling error correction using the viterbi algorithm, edit-distance, and a set of rules respectively. Although, most of the recent and state-of-the-art methods of different languages utilize sequence-to-sequence learning by employing encoder-decoder architecture [7, 27, 29, 31, 32, 35]. Moreover, these recent approaches seem to outperform the rule-based methods by a convincing margin. A sequence-to-sequence character level model, where bidirectional LSTM RNN cells are utilized in both encoder and decoder, has been used for Hindi spelling error correction [7, 27]. Etoori et al. [7] further devoted the same approach to Telugu spelling error rectification. Likewise, [29], [31], [35], and [32] mimic a similar encoder-decoder architecture for Punjabi, Azerbaijani, Sinhala, and Malayalam spelling error correction respectively. Among these methods, [29], [31], and [32] employ LSTM cells in both the encoder and decoder. Recently, Sonnadara et al. [35] presented three different neural spell checkers including a character-level CNN based, a semi-character RNN based, and a nested RNN based method for Sinhala spelling error correction. They achieved the highest performance using the semi-character RNN based method which is the current state-of-the-art method among Sinhala spell checkers.

In this article, we propose DPCSpell, a novel transformer-based framework for spelling error correction, addressing the limitations of existing methods that indiscriminately correct all characters in a word. Our method selectively corrects only the erroneous portion, leading to improved performance and making it state-of-the-art for Bangla spelling error correction. Additionally, we have developed a large-scale parallel corpus for the Bangla SEC and made it publicly accessible, overcoming the lack of a publicly available corpus. Furthermore, we observed that existing methods tend to rely on private corpora and withhold their codes, hindering reproducibility. In the spirit of transparency, we have made all our codes publicly accessible, fostering a reproducible baseline for the task.

## 3 Corpus Creation

Our extensive study found that there could be 14 types of spelling errors in Bangla text [5, 13, 23]. These non-word error types can be classified into five major categories namely phonetic [8, 12, 14, 17], visual [13], typographical [5, 26], run-on [13], and split-word errors [23]. The elucidations for all 14 types of errors are explicated as follows:

- **Phonetic Error:** This phenomenon arises when characters or words with analogous articulation result in different semantic connotations. Consequently, it can be further classified into cognitive and homonym errors, depending on character or word-level factors.

  - **Cognitive Error:** It is a character-level error caused by the similarity in the pronunciation of different letters. There exist several Bangla character clusters, such as ই-ঈ, প-ফ, জ-ঝ-য, ন-ণ, স-শ-ষ and so forth, in which each character has a similar sound. Example: পানি $\rightarrow$ ফানি, অসূয়াপূর্ণ $\rightarrow$ অষূয়াপূর্ণ, পরনির্ভরশীল $\rightarrow$ ফরনির্ভরশীল, etc.

  - **Homonym Error:** It is a word-level error in which multiple words have similar sounds but different meanings and spellings. However, it is context-dependent as the spelling of the erroneous word in a sentence itself could be correct. For example, শিকার—স্বীকার, ক্রীত—কৃত, যোগ্য—যজ্ঞ , etc.

- **Visual Error:** The similar visual shape of different characters causes visual errors which could further be classified into unique character level and combined character level errors.

  - **Unique Character:** It occurs due to the similar shapes of different single characters, such as ই-ঈ, উ-ড-ঊ, ঔ-ঐ, ব-র, etc. Example: যত্নসহকারে $\rightarrow$ যত্নসহকাবে, ডাগরআঁখি $\rightarrow$ ড়াগরআঁখি.

  - **Combined Character:** Unlike unique character level errors, it happens because of the similar visual shape of different combined characters with other combined or single characters, like ত্র-এ, ত্ত-ও, ষ্ক-স্ক, স্ট-ষ্ট, and so on. Example: ছাত্রলীগ $\rightarrow$ ছাএলীগ, অংস্কান্ত $\rightarrow$ অংষ্কান্ত.

- **Typographical Error:** It refers to typing mistakes that occur when we inadvertently press the wrong key while writing. Moreover, it could further be split into the following four categories based on mistakes:

  - **Typo Deletion:** It results from skipping characters while typing a word. For example, চৌচাপটে $\rightarrow$ চৌচাটে, চন্দ্রমুখ $\rightarrow$ চন্দ্রমখ , যত্নবতী $\rightarrow$ যত্নতী , etc.

- **Typo Substitution:** It happens because of pressing the wrong key to write a particular character. Since there exist both phonetic and unicode-based Bangla keyboards, we consider the most popular keyboards of these types, namely Avro and Bijoy. That means typo substitution errors are generated with respect to two keyboards:
    * Typo substitution errors for Avro keyboard
    * Typo substitution errors for Bijoy keyboard

    Example: ঠকান → ফকান(Bijoy Keyboard), টকান(Avro Keyboard).

  - **Typo Transposition:** It emerges as a result of putting the $(N+1)^{th}$ letter of a given word in place of the $N^{th}$ letter and the $N^{th}$ letter in place of the $(N+1)^{th}$ letter, such as বাসভবন → বাসবভন, অলংঙ্কার → অংলঙ্কার, ধুম্বল → ধুল্মব, etc.

  - **Typo Insertion:** It occurs due to the inadvertent inclusion of a character or redundant letter while writing a word. For example, রসাত্মক → রসাতত্মক, উপযোগীকরণ → উপযযোগীকরণ , ভূতের → ভূতেরর, etc.

- **Split-word Error:** It is caused by adding additional space when writing a word which essentially produces two words where one of these three cases could happen: (a) both words are correct, (b) both words are wrong, and (c) only one of them is correct. Hence, it could be classified into the following four types based on where the extra space is placed.

  - **Split-word Both:** The space is placed in such a fashion, case (a), that both sides of the space form two correct words. Example: ক্ষীণচন্দ্র → ক্ষীণ  চন্দ্র, উচ্চবর্ণের → উচ্চ  বর্ণের.

  - **Split-word Random:** It generates two incorrect words because of the inclusion of extra space which denotes case (b). Example: রক্তক্ষয়ী → রক্  তক্ষয়ী, দর্শনযোগ্য → দর্  শনযোগ্য.

  - **Split-word Left:** The space is placed in such a way, case (c), that the left side of the space remains a valid word whereas the right side is not. Example: অসম্পন্নতাবোধক → অসম্পন্ন তাবোধক, প্রতিকূলাচরণ → প্রতি কূলাচরণ.

  - **Split-word Right:** It is the opposite of the split-word left error. Here, the left side of the space turns into an incorrect word whereas the right side becomes a valid word. Example: উপভোগকারী → উপ  ভোগকারী, ইষ্টাপূর্ত → ইষ্টা  পূর্ত.

- **Run-on Error:** It arises from the omission of a space between two correct words in a sentence, resulting in the formation of an incorrect word. Example: রাস্তাঘাট (তৈরি) → রাস্তাঘাটতৈরি, বাংলার (মাটি) → বাংলারমাটি

We create a large-scale parallel corpus for Bangla spelling error detection and correction by incorporating all 14 types of spelling errors. It begins with collecting unerring words, followed by synthetic error generation and further error filtration. Moreover, our proposed method for synthetic error generation is applicable to any Bangla text corpus.

## 3.1 Word Accumulation

The process of synthetic data generation begins with the collection of error-free words through web scraping, which involves automatically extracting web data from a website by parsing its HTML code [37]. In our case, we focus exclusively on the online dictionary domain, which contains accurate words. We gather our data from a popular open-source Bengali-to-Bengali dictionary[1]. To begin, we define a set of $N$ distinct Bengali characters denoted by $C = \{C_1, C_2, ..., C_N\}$, where $C_i$ represents the $i^{th}$ character. We perform web scraping to retrieve raw texts from the dictionary for each character $C_i \in C$. For this purpose, we utilize two well-known web scraping libraries: Requests[2] and BeautifulSoup[3]. Requests is used to extract the HTML codes, while BeautifulSoup helps in extracting the text data. Since the open-source dictionary provides a search engine that allows us to filter out words starting with a specific character $C_i \in C$, we generate the corresponding URLs for each character, where all the words starting with that character can be found. The Requests library is employed to make HTTP requests and retrieve the HTML code response, which we then parse. Using the BeautifulSoup library with the LXML parser, we extract the desired text from the HTML code response. At this stage, we obtain a list of error-free words that require further preprocessing. To facilitate the cleaning process, we construct another set of non-repetitive, frequently occurring Bengali characters, which includes a space, denoted as $D = \{D_1, D_2, ..., D_K\}$. We then iterate through the collected unprocessed text data and remove any characters that are not present in our constructed set of frequent Bengali characters, $D$. Finally, all the preprocessed errorless words, represented as $W = [W_1, W_2, ..., W_p]$, where $W_k \in W$ is an error-free word starting with $C_i$, are stored in a CSV file and saved on the local machine for further use.

## 3.2 Error Annexation

At this stage, we introduce the non-word errors discussed earlier into the errorless words list $W$. Separate dictionaries are created to introduce cognitive, visual, typographical substitution, and run-on errors. These dictionaries can be represented as $\{P_1 : L_1, P_2 : L_2, ..., P_N : L_N\}$, where $P_n$ represents a key such that $P_n \in C$, and $L_n$ is the list of potential erroneous characters associated with that key $P_n$. The dictionaries consist of $N$ character keys denoted as $P = [P_1, P_2, ..., P_N]$, where $P_n$ is the $n^{th}$ character in $C$. Additionally, $L_n$ is the list of potential erroneous characters for each key $P_n \in P$, denoted as $L_n = [L_1, L_2, ..., L_R]$, where $L_i$ represents the $i^{th}$ potential erroneous character such that $L_i \in D$ and $L_i \neq P_n$. To introduce cognitive, visual, typographical, substitution, and run-on errors to the words $W_i \in W$, we make use of dictionaries that were previously constructed. However, we employ list comprehension to introduce typographical deletion, transposition, insertion, and split-word errors. Additionally, we compile a list of homonym errors using a similar approach employed for gathering the list of error-free words, $W$. It is worth noting that we deliberately modify a single character of an accurate word to introduce different types of errors. Ultimately, we construct a large-scale parallel corpus denoted as $M = \{S_1 : T_1, S_2 : T_2, ..., S_t : T_t\}$, where each $M_i$ represents an $i^{th}$ source $(S_i)$ - target $(T_i)$ pair, with the source being the erroneous word and the target being the accurate word.

[1] https://accessibledictionary.gov.bd/
[2] https://pypi.org/project/requests/
[3] https://pypi.org/project/beautifulsoup4/

### 3.3 Error Filtration

Several types of synthesized errors, namely typo deletion, typo Avro substitution, and typo Bijoy substitution, appear to generate certain vague errors to some extent. These errors are produced by an employed dictionary lookup table ($\{P_1 : L_1, P_2 : L_2, ..., P_N : L_N\}$), resulting in a wide range of error variations, surpassing the typical range of human mistakes. To address this, we employ a transformer-based language model depicted in Figure 1(a) to identify and filter out these rare and unnecessary errors, thereby enhancing the sophistication of the corpus. The error filtration model effectively removes these infrequent errors. We utilize the same transformer architecture employed in the detector, purificator, and corrector networks for error filtration. By incorporating a modified corrector network in our DPCSpell, we exclude the atypical error patterns associated with these three error types that are unlikely to be made by humans. We have made the final corpus publicly available, which can be found at https://tinyurl.com/DPCSpell.

| **Error Type** | **#No. of Instances** | **Percentage** |
|---|---|---|
| Cognitive Error | 186,620 | 13.52% |
| Homonym Error | 123 | 0.01% |
| Visual Error (Single Character) | 113,912 | 8.25% |
| Visual Error (Combined Character) | 17,313 | 1.25% |
| Typographical Deletion | 102,550 | 7.43% |
| Typographical Substitution (Bijoy) | 222,930 | 16.15% |
| Typographical Substitution (Avro) | 174,248 | 12.62% |
| Typographical Transposition | 122,939 | 8.90% |
| Typographical Insertion | 124,767 | 9.04% |
| Run-on Error | 124,895 | 9.05% |
| Split-word Error (Left) | 51,610 | 3.74% |
| Split-word Error (Right) | 13,985 | 1.01% |
| Split-word Error (Random) | 111,974 | 8.11% |
| Split-word Error (both) | 12,798 | 0.93% |
| | **Total = 1,380,664** | |

Table 1: Statistic of the Bangla SEC corpus

### 3.4 Corpus Statistic and Error Percentage Validation

Table 1 provides an overview of the statistics for the final corpus. Among the different error types, typographical substitution (Bijoy) exhibits the highest number of erroneous pairs, comprising 16.15% of the corpus, followed by typographical substitution (Avro) at 12.62%. In contrast, homonym errors account for a mere 123 instances, representing the smallest proportion among all error types. Examining individual error categories, five out of the 14 categories contain less than 5% errors, while typographical substitution (Bijoy) stands out with over 15% errors. The remaining categories exhibit approximately 10% errors each. Among the five major error categories, phonetic, visual, typographical, run-on, and split-word errors account for 13.53%, 9.50%, 54.14%, 9.05%, and 13.77% of the corpus, respectively. It is noteworthy that typographical errors contribute slightly more than half of the total errors, aligning with our initial expectations based on inspection.

The percentages of errors in the corpus are not predetermined by us, as they depend on the characteristics of the corpus itself. Specifically, the error annexation process relies solely on the constructed dictionaries for each error type. Considering the potential set of erroneous characters ($L_n \in L$) for each key in the dictionary ($P_n \in P$), it is expected that the number of typographical errors would be higher compared to other error types. To validate the error percentages in the corpus, we conducted a thorough analysis of human tendencies to make errors while writing, using real-world data. We collected a substantial amount of data from the comment sections of public Facebook[4] posts. Subsequently, we cleaned the corpus of these comments and searched for erroneous words based on our generated corpus ($M$). The results of this inspection provided further validation of the percentage distribution of different error types in our corpus. In most categories, we found a similar proportion of erroneous words in the real-world corpus, confirming the accuracy of our corpus composition. It is worth noting that the percentage of typographical deletion errors deviated from the expected proportion, which was anticipated due to the reasons mentioned earlier. However, to ensure fairness in the inspection process, we created a balanced version of our corpus where no error type accounted for more than 10% of the entire corpus. This approach allows for a comprehensive evaluation of the different error categories within a controlled and equitable framework.

# 4 Methodology

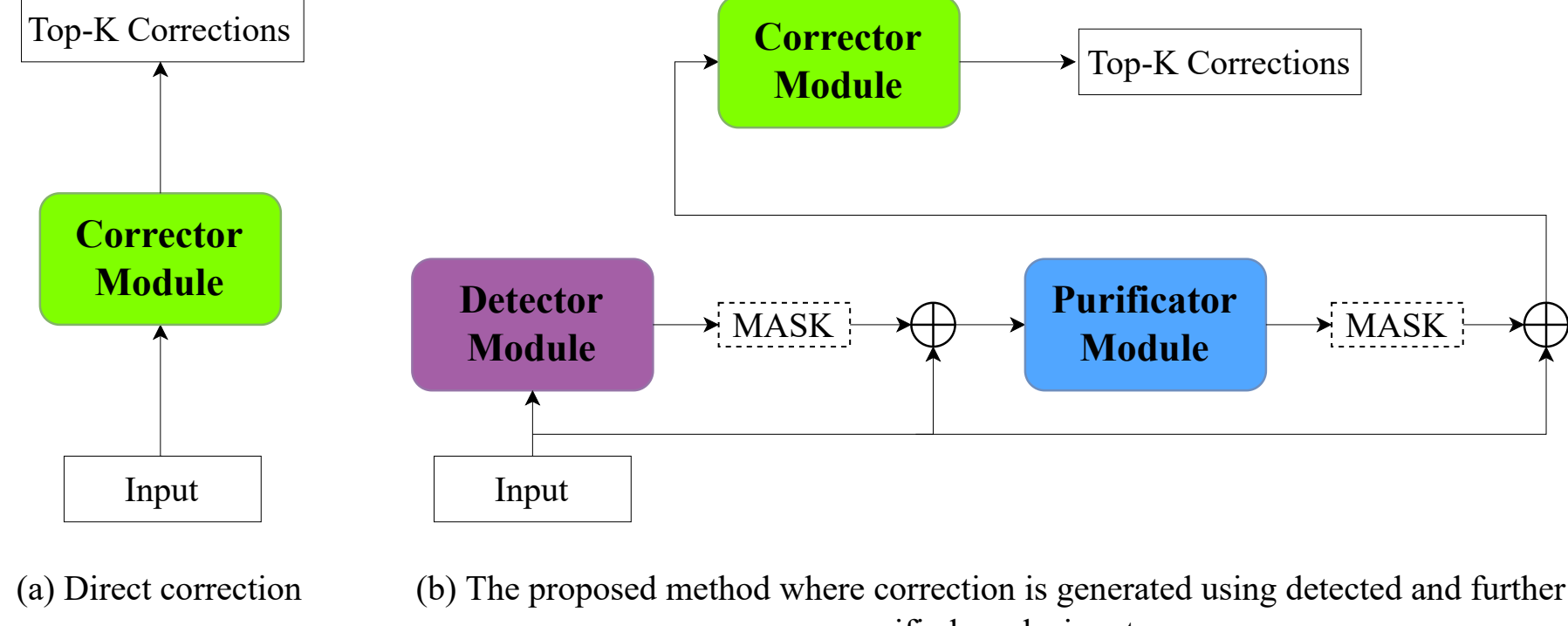

(a) Direct correction

(b) The proposed method where correction is generated using detected and further purified masks input

Figure 1: **(Left)** A direct correction approach which takes a sequence as input and corrects the whole sequence regardless of its correctness. **(Right)** Our proposed DPCSpell which takes a sequence as input and detects and purifies the erroneous portion in the sequence. Finally the corrector module fixes the faulty segment of the sequence only.

[4] https://www.facebook.com/

## 4.1 Problem Formulation

The character-level spelling error rectification task aims to map an erroneous sequence denoted as $X_I = \{X_{I_1}, X_{I_2}, ..., X_{I_N}\}$ into the corresponding correct sequence represented as $\hat{Y} = \{\hat{Y_1}, \hat{Y_2}, ..., \hat{Y_M}\}$ where $X_{I_j}$ and $\hat{Y_j}$ are characters of the same language, and $N \in Z^+$ and $M \in Z^+$ but not necessarily required to be equal.

Our proposed method consists of three modules: the detector module $\mathcal{D}(.)$, the purificator module $\mathcal{P}(.)$, and the corrector module $\mathcal{C}(.)$. In the detector module, $X_I$ is inputted to identify the positions of erroneous characters. These erroneous characters are then replaced with a special token, $[MASK]$, resulting in $X_D = \{X_{D_1}, X_{D_2}, ..., X_{D_N}\}$, where $X_{D_j}$ is equal to $X_{I_j}$ if the $j^{th}$ character is correct, otherwise it is the special token, $[MASK]$. Next, in the purificator module, an amalgamation of $X_I$ and $X_D$, denoted as $X_{ID} = \{< SEP > +X_I + < SEP > +X_D + < SEP >\}$, is used as input. The special token $< SEP >$ is used to distinguish between $X_I$ and $X_D$. The purificator module further refines the detected erroneous positions, resulting in $X_P = \{X_{P_1}, X_{P_2}, ..., X_{P_N}\}$. Finally, the corrector module focuses on correcting only the detected erroneous characters rather than correcting all the characters in $X_I$. It combines the initial erroneous sequence $X_I$ with the detected and purified positions of erroneous characters in $X_P$. This combined sequence is represented as $X_{IP} = \{<SEP> +X_I+ <SEP> +X_P+ <SEP>\}$. The $X_{IP}$ is then used as input to the model, which generates the corresponding correct sequence $\hat{Y}$.

## 4.2 Overview of DPCSpell

The working mechanism of DPCSpell, depicted in Figure 2, is an amalgamation of a detector, purificator, and corrector network. The detector network ($\mathcal{D}(.)$) employs a transformer to identify the positions of incorrect characters ($X_D$) in an input sequence $X_I$. Similarly, both the purificator ($\mathcal{P}(.)$) and corrector ($\mathcal{C}(.)$) networks utilize transformers for their respective tasks. The purificator takes an amalgamation of $X_I$ and $X_D$ as input and produces a more precise mask, $X_{ID}$. Finally, the corrector transformer utilizes $X_I$ and $X_{ID}$ to generate the correction $\hat{Y}$. To summarize, the framework's detector module receives an erroneous word as input and attempts to detect the erroneous characters. On the other hand, the corrector module utilizes the refined mask from the purificator module, which incorporates the output of the detector, to generate appropriate corrections. The entire procedure can be mathematically represented as follows:

$$\hat{Y} = \mathcal{C}((X_I^T, \mathcal{P}((X_I^T, \mathcal{D}(X_I^T; W_D)); W_P)); W_C) \tag{1}$$

## 4.3 Motivations

The recent emergence of NLP and deep learning has achieved astonishing success in spelling error correction. Currently, state-of-the-art spell checkers of different resource-constrained languages leverage seq2seq models and employ a direct correction approach, as exemplified in Figure 1(a). However, these methods are end-to-end in nature and exhibit enormous false alarm rates, as they correct all the characters of the sequence regardless of its correctness. This problem is exacerbated when just a tiny percentage of wrong characters occur in the entire sequence, which appears to be a general trend, as evidenced in Table 2.

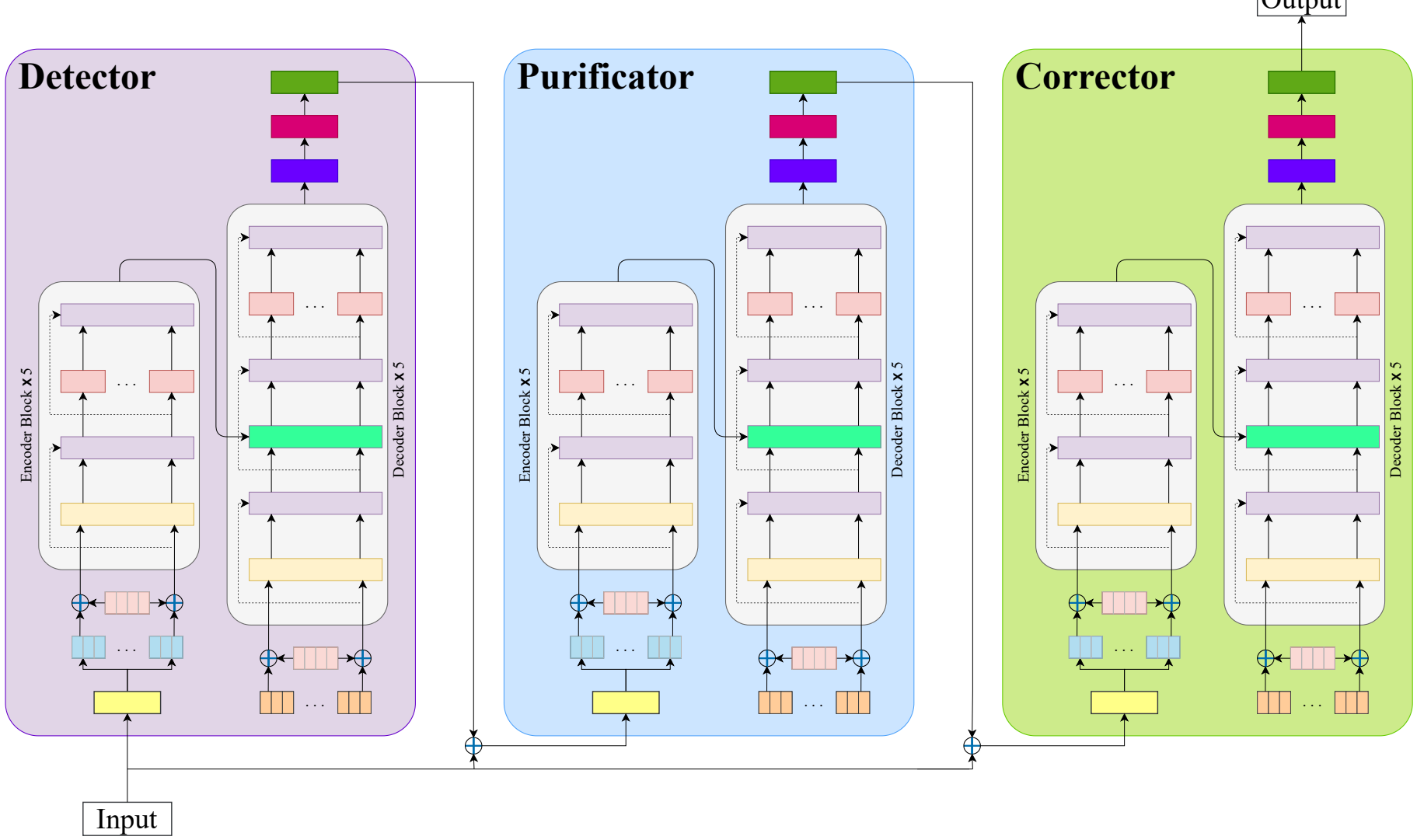

Figure 2: **(Left)** The detector network of DPCSpell which takes an input sequence as input and makes an initial attempt to detect the erroneous portions of the sequence. **(Middle)** The purificator network which takes the actual input sequence along with the detected sequence from the detector to further purify the erroneous fragments of the sequence. **(Right)** The corrector network which combines actual input with the purified sequence from the purificator to generate the correction.

| **Method** | **Input** | **Output** |
|---|---|---|
| Direct | অফরিনত → অ ফ র ◌ি ন ত | অ প র ◌ি ণ ত → অপরিণত |
| DPCSpell | অফরিনত → অ ফ র ◌ি ন ত | অ প র ◌ি ণ ত → অপরিণত |

Table 2: Comparison between the working mechanism of direct approaches and our proposed DPCSpell

DCSpell [38] addresses this problem by presenting a transformer-based detector-corrector framework that first determines whether a character is erroneous or not before correcting it, which we found rather inefficacious in correction generation. Since the correction generation largely relies on the detected erroneous character sequence, it often fails to precisely identify the positions of erroneous characters which essentially misguides the corrector network. We resolve this by introducing a purificator in between the detector and corrector module that further cleanses the detected erroneous characters. Our proposed DPCSpell eliminates the drawbacks associated with direct correction approaches as it only corrects the wrong characters in the sequence by detecting the erroneous portions in the input sequence beforehand. The addition of the purificator module delineates a lucid improvement over the detector-corrector framework. Furthermore, DPCSpell converges much quicker than both the detector-corrector and direct correction strategies. Figure 1(b) depicts our proposed DPCSpell, consisting of the detector, purificator, and corrector modules.

## 4.4 Structure of DPCSpell

The building blocks of the transformer used in the detector, purificator, and corrector network is delineated in Figure 3. The encoder and decoder of the transformer comprised of a stack of five residual encoding and decoding blocks respectively.

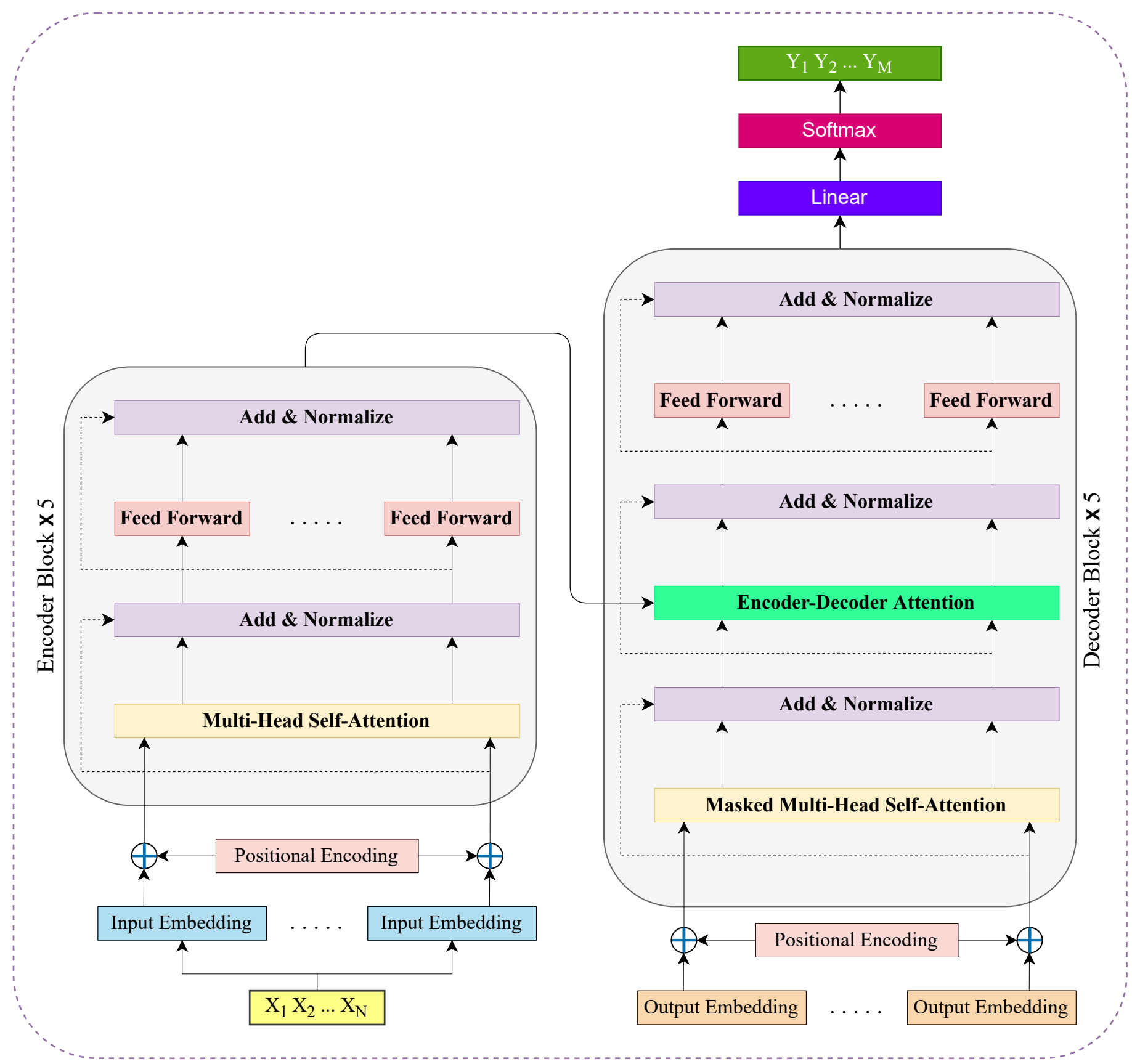

Figure 3: **(Left)** The encoding component of the transformer is a stack of identical encoders, which is responsible for mapping an input sequence to a sequence of dense vector representations. **(Right)** The decoding module which is also a stack of identical decoders. It accepts the encoded output from the encoder along with the decoder output from the previous time step to generate a prediction sequence.

**Encoder:** The encoder takes the input sequence, $X_I = \{X_{I_1}, X_{I_2}, ..., X_{I_N}\}$, and compresses it into a sequence of context vectors, $Z = \{Z_1, Z_2, ..., Z_K\}$, which capture the information from all tokens in $X_I$, including their positions. To achieve this, each token $X_{I_i} \in X_I$ is first passed through an embedding layer, and then the positional encoding is added elementwise. This results in integrated embeddings that are ready to be processed by the multi-head attention layer. The multi-head attention layer incorporates $h$ different attentions calculated from $h$ heads in parallel. The self-attention can be seen as a combination of queries, keys, and values, where a scalar self-attention score is calculated by taking the dot product between the query and the key. The score is then divided by the square root of the size of the key vector and passed through a softmax layer. The resulting softmax score is then multiplied by the value vector, generating an attention vector for a specific head. The formula for calculating the multi-head attention is as follows [39].

$$MultiHead(Q, K, V) = Concat(Head_1, Head_2, ..., Head_h)W^o \tag{2}$$

$$Head_i = Attention(QW_i^Q, KW_i^K, VW_i^V) \tag{3}$$

$$Attention(Q, K, V) = Softmax(\frac{QK^T}{\sqrt{d_k}})V \tag{4}$$

Here, $Q$, $K$, and $V$ refer to query, key, and value respectively and $W_i^j$ is their corresponding weight matrix. And $d_k$ denotes the dimension of the key vectors.

Next, the responses of multi-head attention layer is then passed through position-wise feed-forward neural networks incorporating residual connections from input of the layer. The output of feed-forward layer is further normalized which is then ready to pass through next encoder block, except for the final encoding block which outputs a sequence of context vectors $Z = \{Z_1, Z_2, ..., Z_N\}$ for the decoder.

**Decoder:** The decoder ($\delta$) is analogous to the encoder ($\xi$) but with two distinct multi-head attention layers: self-attention (masked) and encoder-attention. In the self-attention layer, the decoder representation serves as the query, while the encoder representations are used as the key and value in the standard multi-head attention layer for encoder-attention. The decoder representation from the final decoding block is then fed through a linear layer, followed by a softmax activation function, generating the output sequence $Y = \{Y_1, Y_2, ..., Y_M\}$.

### 4.4.1 Detector Network

Given an input sequence $X_I = \{X_{I_1}, X_{I_2}, X_{I_N}\}$, the detector aims to identify and label potential erroneous characters. It returns a labeled sequence $X_D = \mathcal{D}(X_I) = \{X_{D_1}, X_{D_2}, X_{D_N}\}$, where the positions of potential erroneous characters are masked. If the $i^{th}$ character $X_{I_i} \in X_I$ is deemed erroneous, it is replaced with a special token called [MASK] ($X_{D_i} = [MASK]$). Otherwise, it copies the corresponding character from the input sequence ($X_{D_i} = X_{I_i}$, as shown in eq. 5). The detected erroneous sequence $X_D$ is generated by passing the decoded representation of $X_I$ through a linear layer, $\mathcal{F}_d$.

$$X_{D_i} = \begin{cases} [MASK], & \text{if } softmax(\mathcal{F}_d(\delta_d(\xi_d(X_{I_i})) = index([MASK]) \\ X_{I_i}, & \text{otherwise} \end{cases} \tag{5}$$

We use the categorical cross-entropy loss function to measure the goodness of fit of our method which is demonstrated in eq. 6 and adam optimizer with a constant learning rate to minimize the loss.

$$CE_{\mathcal{D}} = -\sum_{i}^{N} X_{D_{actual}} log(\mathcal{D}(softmax(\mathcal{F}_d(\delta_d(\xi_d(X_{I_i})))) \tag{6}$$

Where $X_{D_{actual}}$ is the ground truth sequence of $X_I$ and $\mathcal{D}(softmax(\mathcal{F}_d(\delta_d(\xi_d(X_{I_i})))$ is the detected sequence from the detector, $\mathcal{D}$.

#### 4.4.2 Purificator Network

The purificator module is identical to the detector except for the input sequence. It concatenates the detected sequence $X_D$ with initial input $X_I$ such that $X_{ID} = \{<\text{SEP}> +X_I+ <\text{SEP}> +X_D+ <\text{SEP}>\}$. It further purifies the detected erroneous character sequence $X_D$ and returns an updated labeled sequence $X_P = \mathcal{P}(X_I, X_D) = \{X_{D_1}, X_{D_2}, ..., X_{D_N}\}$ from $X_{ID}$. Mathematically $X_P$ can be defined as follows:

$$X_{P_i} = \begin{cases} [MASK], & \text{if } softmax(\mathcal{F}_p(\delta_p(\xi_p(X_{I_i}, X_{D_i})) = index([MASK]) \\ X_{I_i}, & \text{otherwise} \end{cases} \tag{7}$$

Similar to detector, here we use cross-entrophy loss function and adam optimizer with an unvarying learning rate.

$$CE_{\mathcal{P}} = -\sum_{i}^{N} X_{D_{actual}} log(\mathcal{P}(softmax(\mathcal{F}_p(\delta_p(\xi_p(X_{I_i}, X_D)))) \tag{8}$$

#### 4.4.3 Corrector Network

The corrector module takes an amalgamation of the initial input sequence $X_I$, as well as the detected and further purified masks $X_P$ from the purificator, as input. It generates the corrections $\hat{Y} = \mathcal{C}(X_I, X_{IP}) = \{\hat{Y}_1, \hat{Y}_2, ..., \hat{Y}_N\}$. Similar to the purificator, it concatenates $X_I$ and $X_P$ in the form of $X_{IP} = \{<\text{SEP}> +X_I+ <\text{SEP}> +X_P+ <\text{SEP}>\}$. The vocabulary of corrector is denoted as $V = \{V_1, V_2, ..., V_N\}$ where $V_i$ is the $i^{th}$ character of a particular language. For each character position $X_{IP_i} \in X_{IP}$ with the $[MASK]$ token, the corrector ($\mathcal{C}(.)$) predicts a character $\hat{Y}_i \in V$. The encoder ($\xi_c(.)$) of the corrector network generates a sequence of context vectors $Z_{\mathcal{C}} = \{Z_{\mathcal{C}_1}, Z_{\mathcal{C}_2}, ...Z_{\mathcal{C}_N}\}$ for each character $Z_{IP_i} \in Z_{IP}$. The encoder representation (ER) is then passed through the decoder ($\delta_c(.)$), and the decoder representation (DR) is processed through a fully connected layer $\mathcal{F}_c$ to generate the correction $\hat{Y}$.

$$ER = \xi_c(X_{IP}) \tag{9}$$

$$DR = \delta_c(ER) \tag{10}$$

$$\hat{Y} = softmax(\mathcal{F}(DR; EN)) \tag{11}$$

Here we employ the cross-entropy loss function to measure the goodness of fit of our corrector network which is denoted in equation-12 and adam optimizer with a steady learning rate to minimize the loss.

$$CE_{\mathcal{C}} = -\sum_{i}^{N} X_{I_{actual}} log(softmax(\mathcal{F}_c(\delta_c(\xi_c(X_{I_i}, X_P)) \tag{12}$$

Where $X_{I_{actual}}$ is the gold standard annotation of $X_I$ and $softmax(\mathcal{F}_c(\delta_c(\xi_c(X_{I_i}, X_P)$ denotes the generated correction from the corrector network $\mathcal{C}$.

## 5 Experimental Analysis

We adopt a standard train-validation-test set approach, widely used for the spelling error correction task [40, 41, 42]. By dividing the data into three sets - training, validation, and test - we prevent the model from merely memorizing the training examples, enabling a better assessment of its ability to generalize. The segregation of these sets is paramount, guaranteeing a dependable appraisal of the model's prowess on previously unseen data. Moreover, when comparing different models, using the same test set is pivotal to ensure fair and accurate comparisons. This rigorous methodology enhances the reliability of our spelling error correction method.

## 5.1 Datasets

### 5.1.1 Bangla

We use our large-scale parallel corpus for Bangla spelling error correction. To do so, we split our corpus into training, validation, and test sets for further use. However, the instances of our Bangla SEC corpus have been exemplified in Table 3.

**Training Set** The training set accounts for 80% of the data in the corpus. We take 80% of the data from each individual error type to prevent the corpus from being biased. The training set comprises 1,104,531 correct-erroneous word pairs.

**Validation Set** We keep only 5% of the data from the corpus in the validation set. Similar to the training set, we consider 5% of each error category to construct this set as well. As a result, the validation set contains 69,034 instances.

**Test Set** It is comprised of 15% errors of all 14 error types, as we did in the training and validation sets. It accounts for 207,099 instances of the corpus.

| **Source** | **Mask** | **Target** | **Error Type** |
|---|---|---|---|
| জাগরন | জাগর_ | জাগরণ | Cognitive Error |
| উদ্ধত | উদ্_ত | উদ্যত | Homonym Error |
| শোভাকবণ | শোভাক_ণ | শোভাকরণ | Visual Error (Single Character) |
| এইক্ষ্মণ | এই__ণ | এইক্ষণ | Visual Error (Combined Character) |
| ভমিষ্ট | _মিষ্ট | ভূমিষ্ট | Typographical Deletion |
| যোহ্যতর | যো_্যতর | যোগ্যতর | Typographical Substitution (Bijoy) |
| ঈষৎভঞ্চল | ঈষৎ_ঞ্চল | ঈষৎচঞ্চল | Typographical Substitution (Avro) |
| নবড়ড়ে | নড়বড়ে | ন__ড়ে | Typographical Transposition |
| উপযযোজন | উপয_োজন | উপযোজন | Typographical Insertion |
| জাগ্রৎকুর্ম | জাগ্রৎ_____ | জাগ্রৎ | Run-on Error |
| জন্ম চক্রে | জন্ম_চক্রে | জন্মচক্রে | Split-word Error (Left) |
| চিহ্ন বিশেষ | চিহ্ন_বিশেষ | চিহ্নবিশেষ | Split-word Error (Right) |
| মহৌ ষধ | মহৌ_ষধ | মহৌষধ | Split-word Error (Random) |
| অস্থি বিষয়ক | অস্থি_বিষয়ক | অস্থিবিষয়ক | Split-word Error (both) |

Table 3: Instances from the Bangla SEC corpus

### 5.1.2 Hindi and Telugu

We utilize the Hindi and Telugu parallel corpora used in [7] along with their training and test sets. The training and test sets of the Hindi corpus contain 90,489 and 9,049 instances, respectively. Likewise, there are 64,518 training and 7,727 test pairs in the Telugu corpus.

**Hindi*** We enhance the corpus by introducing nine types of errors including cognitive, visual (single), typographical insertion, typographical deletion, typographical transposition, run-on, split-word left, split-word right, split-word random, and split-word both errors. The enriched training and test sets include 177,038 and 19,660 instances, respectively.

**Telugu*** We bring forward variety in the Telugu corpus by incorporating those nine errors that were previously introduced in the Hindi corpus. Consequently, the enhanced training and test set contain 214,828 and 20,279 correct-erroneous word pairs, respectively.

| **Hindi** | | | **Telugu** | | |
|---|---|---|---|---|---|
| **Source** | **Mask** | **Target** | **Source** | **Mask** | **Target** |
| अनियमीत | अनियम_त | अनियमित | మొ్తం | మొ_తం | మొత్తం |
| साक | स_क | सक | జ్యోతి లక్ష్మి | జ్యోతి_లక్ష్మి | జ్యోతిలక్ష్మి |
| अनुष्ठाना | अनुष्ठान_ | अनुष्ठान | స్థానాు | స్థానా_ | స్థానాలు |

Table 4: Examples from the enhanced Hindi **(left)** and Telugu **(right)** SEC corpus

## 5.2 Baselines

We compare our method with seven baselines including several state-of-the-art methods of different resource-scarce languages.

- **RuleBased [13]:** This method utilizes Double Metaphone and Edit Distance algorithms for Bangla spelling error detection and correction.

- **GRUSeq2Seq:** Bahdanau et al. [43] enriches the conventional RNN encoder-decoder architecture, by allowing the model to focus only on the pertinent details from the encoder while generating a target word, for neural machine translation, where GRU is employed in both the encoder and decoder. We make it a baseline for Bangla spelling error correction through Bangla-to-Bangla translation.

- **LSTMSeq2Seq [7]:** This method brings forward a character level seq2seq model utilizing LSTM cells in both the encoder and decoder for spelling error correction of two resource-scarce Indic languages namely Hindi and Telugu.

- **ConvSeq2Seq:** Gehring et al. [44] presents a fully convolutional sequence-to-sequence architecture with an attention module for neural machine translation. We consider it as another baseline to rectify Bangla spelling errors through Bangla-to-Bangla translation.

- **VocabLearner [2]:** This method introduces a word-level vocabulary learner for Bangla spelling error correction by employing a 1D CNN-based architecture named Coordinated CNN (CoCNN).
- **DTransformer [45]:** This method utilizes a denoising autoencoder transformer for spelling error correction, on a short input string, for four resource-limited languages. The autoencoder is employed for synthetic error annexation, whereas the transformer is responsible for error rectification.
- **DCSpell [38]:** This method initiates a transformer-based detector-corrector framework, where a character is detected first whether it is erroneous or not before being corrected, to rectify Chinese spelling errors.

## 5.3 Performance Evaluation

We evaluate the performance of our method using Precision, Recall, F-scores, Exact Match, and Modified Accuracy.

### 5.3.1 Precision, Recall, and F$\beta$-score

Precision denotes the credibility of a model by signifying the quality of its positive predictions, whereas recall quantifies the proportion of actual positives precisely identified by the model. Precision is beneficial in such situations when a False Positive (FP) is more of a concern than a False Negative (FN). In contrast, recall is a useful metric in such scenarios where False Negative (FN) is highly expensive. The formulas for calculating the precision and recall are as follows.

$$Precision = \frac{\sum_{i=1}^{n} |g_i \cup e_i|}{\sum_{i=1}^{n} |e_i|} \tag{13}$$

$$Recall = \frac{\sum_{i=1}^{n} |g_i \cup e_i|}{\sum_{i=1}^{n} |g_i|} \tag{14}$$

where $g_i$ and $e_i$ denote gold-standard targets and model's predicted levels for $i^{th}$ word such that $W_i \in W$. The intersection for gold-standard targets and model's predicted levels for a given word $M_i \in M$ is considered as,

$$g_i \cup e_i = \{e \in e_i \mid \exists g \in g_i, match(g, e)\} \tag{15}$$

F-measure is the harmonic mean of precision and recall. It is required for comparing different models with high recall and low precision scores. We calculate the F$\beta$-scores for $\beta$ values of 1 and 0.5. The formula for calculating F$\beta$-score is as follows.

$$F\beta\ score = \frac{(1 + \beta^2) \times Precision \times Recall}{(\beta^2 \times Precision) + Recall} \tag{16}$$

***Eq. 16:*** *$\beta$=1 (F1-score) and $\beta$=0.5 (F0.5-score) denote equal weighting of precision and recall, and emphasize on precision while calculating the score*

### 5.3.2 Exact Match (EM)

It delineates the efficacy of the model across all classes like accuracy ($= (TP + TN)/(TP + FP + TN + FN)$), where TP, TN, FP, FN refers to True Positive, True Negative, False Positive, and False Negative). The output of the model ($\mu(x)$) is deemed to be correct when the prediction ($\hat{y}$) exactly matches the label ($y$). The equation is as follows.

$$f(x) = \begin{cases} 1, & \text{if } \hat{y} = \mu(x) = y \\ 0, & \text{otherwise} \end{cases} \tag{17}$$

EM is the ratio of the number of correct predictions and total instances. The higher the EM score, the better the model performance. The formula for calculating the EM score is a follows.

$$EM = \frac{\sum_1^N f(x)}{N} \tag{18}$$

***Eq. 18:*** *where* $\sum_1^N f(x)$ *is the number of correct prediction, and* $N$ *refers to the number of instances*

### 5.3.3 Modified Accuracy (MA)

We calculate the accuracy within the top-K predictions and call it Modified Accuracy. Unlike accuracy, in our case Exact Match, it elucidates the effectiveness of a model over corpora. The prediction is considered positive if any outcome within top-K can be found in the desired corpus. The formula to evaluate a prediction whether it is positive or not is as follows.

$$g(x) = \begin{cases} 1, & \text{if } \hat{y} = \mu(x) = topK \; \epsilon \; W \\ 0, & \text{otherwise} \end{cases} \tag{19}$$

MA, similar to EM, is calculated as the ratio of total positive predictions and instances of the corpus. A higher MA score denotes the credible performance of the model. The formula for calculating the MA score is a follows.

$$MA = \frac{\sum_1^N g(x)}{N} \tag{20}$$

***Eq. 20:*** *where* $\sum_1^N g(x)$ *and* $N$ *denote number of positive predictions and instances in the corpus*

## 5.4 Hyperparameters

The encoder and decoder of our detector, purificator, and corrector network is a combination of 5 encoding and decoding layers respectively. Moreover, we use 8 attention heads in both encoder and decoder. A hidden size of 128 is employed in the encoder and decoder while we kept the pf dimension two-fold of the hidden dimension. Next, a dropout ratio of 10% has been employed in the encoder and decoder in all three modules to avoid overfitting issues. Likewise, we clip the gradient at 1 to eliminate the drawback of exploding gradient. Finally, we use a constant learning rate of $5e - 4$ in adam optimizer to minimize the loss and train the detector, purificator, and corrector network for 100 epochs respectively.

## 5.5 Main Results

### 5.5.1 Spelling Error Correction for Bangla Language

We compare the performances of several state-of-the-art methods and our constructed baselines with our proposed DPCSpell for rectifying Bangla spelling errors. To ensure a fair comparison, we train, validate, and test these methods on our parallel corpus. The empirical outcome of these approaches can be found in table 5.

| Method | EM | MA | PR | RE | F1 | F0.5 |
|---|---|---|---|---|---|---|
| RuleBased [13] | 55.71% | – | 0.5620 | 0.5571 | 0.5578 | 0.5598 |
| GRUSeq2Seq [43] | 75.56% | 76.56% | 0.8072 | 0.7556 | 0.7726 | 0.7899 |
| ConvSeq2Seq [44] | 78.85% | 80.10% | 0.8452 | 0.7885 | 0.8259 | 0.8259 |
| VocabLearner [2] | 22.47% | – | – | – | – | – |
| DTransformer [45] | 90.44% | 91.12% | 0.9061 | 0.9044 | 0.9047 | 0.9056 |
| DCSpell [38] | 84.23% | 85.07% | 0.8458 | 0.8423 | 0.8434 | 0.8446 |
| **DPCSpell** | **94.78%** | **95.16%** | **0.9487** | **0.9478** | **0.948** | **0.9483** |

Table 5: The comparison of the quantitative outcomes of our proposed DPCSpell with other methods in the Bangla SEC task

Our proposed DPCSpell outperforms all the listed methods in table 5 by a convincing margin. It outperforms RuleBase[13], GRUSeq2Seq[43], and ConvSeq2Seq[44] by a higher Exact Match (EM) score of 39.07%, 19.22%, and 15.93%, respectively. Likewise, it outperforms DCSpell[38] by an EM score of 10.55%, a Modified Accuracy (MA) score of 10.09%, a precision (PR) score of 0.1029, a recall (RE) score of 0.1055, an F1 score of 0.1046, and an F0.5 score of 0.1037. Besides, it suppresses the effectiveness of the recent Bangla spelling error correction method named VocabLearner[2] by accomplishing a 72.31% higher EM score. Moreover, it improves the performance of DTransformer[45], which is the second best method to ours, by attaining higher EM, MA, PR, RE, F1, and F0.5 scores of 4.43%, 4.04%, 4.26%, 4.34%, 4.33%, and 4.27%, respectively.

In addition, table 5 depicts a thorough comparison between rule-base, GRU-base, convolution-based, and transformer-based methods. The rule-based method performs the worst followed by GRU-based and convolution-based methods. However, convolution-based method slightly improved the performance of GRU-based methods. In contrast, transformer-based methods show promising result. We compare, one-stage, two-stage, and three-stage transformer-based methods which are DTransformer, DCSpell, and DPCSpell, respectively. The empirical outcome delineates that two-stage DCSpell performs worst among these three, where our proposed three-stage DPCSpell performs the best.

### 5.5.2 Bangla Spelling Error Analysis

To examine the effectiveness of our method in Bangla spelling error correction, we compare the performance of individual error types with two competitive methods in table 6. It outperforms both DTransformer and DCSpell by a significant EM score in all individual error types. Likewise, it exceeds the listed methods in terms of MA in all individual error categories, except homonym error where DTransformer achieves the highest score. However, it performs poorly in the case of correcting homonym errors due to the insufficient number of training instances.

| **Error Type** | DTransformer | | DCSpell | | DPCSpell | |
|---|---|---|---|---|---|---|
| | **EM** | **MA** | **EM** | **MA** | **EM** | **MA** |
| Homonym Error | 11.38% | **73.98%** | 11.38% | 67.04% | **17.07%** | 72.36% |
| Typo Deletion | 90.92% | 91.36% | 79.59% | 80.02% | **94.07%** | **94.38%** |
| Typo Substituition (Bijoy) | 91.80% | 91.95% | 85.98% | 86.26% | **95.55%** | **95.67%** |
| Typo Substituition (Avro) | 93.35% | 93.50% | 89.40% | 90.09% | **97.55%** | **97.63%** |
| Visual Error (Single) | 79.98% | 80.96% | 77.48% | 78.38% | **90.85%** | **91.61%** |
| Cognitive Error | 90.21% | 90.73% | 83.12% | 83.58% | **94.79%** | **95.25%** |
| Typo Transposition | 87.43% | 89.01% | 81.58% | 82.32% | **93.35%** | **94.70%** |
| Visual Error (Combined) | 93.27% | 82.14% | 88.50% | 76.97% | **96.21%** | **82.13%** |
| Run-on Error | 88.32% | 89.37% | 80.71% | 83.34% | **90.09%** | **90.75%** |
| Typo Insertion | 97.16% | 97.27% | 91.66% | 92.73% | **99.68%** | **99.72%** |
| Split-word Error (Left) | 92.64% | 94.63% | 86.65% | 88.13% | **95.38%** | **97.64%** |
| Split-word Error (Right) | 93.88% | 95.26% | 80.00% | 81.09% | **97.71%** | **98.43%** |
| Split-word Error (Random) | 89.80% | 92.74% | 82.85% | 85.29% | **93.96%** | **95.12%** |
| Split-word Error (both) | 93.59% | 95.10% | 86.32% | 87.64% | **96.41%** | **97.36%** |
| **Weighted Average** | 90.44% | 91.12% | 84.23% | 85.07% | **94.78%** | **95.16%** |

Table 6: The comparison of the quantitative outcomes of our proposed DPCSpell in individual error types of the Bangla SEC with other competitive methods

For further analysis, we compare some rectification findings of these methods on the test data in table 7 where the tick and cross marks denote whether the prediction of the method is correct or not. We randomly choose six out of 14 error types to demonstrate the qualitative outcomes of DTransformer, DCSpell, and DPCSpell. The empirical outcomes of our DPCSpell validate its effectiveness in correcting Bangla spelling errors. Especially, our method is proficient in rectifying all types of errors, whereas DTransformer and DCSpell suffer from correcting words with longer sequences and combined characters. However, the mistakes made by our method are also quite relevant. In the case of homonym errors, even though it fails to generate the actual correction (আহুতি) of the erroneous word (আহূতি), the prediction (আহত) itself is a correct word and makes complete sense. In another example, our DPCSpell generates কর্কশ for the erroneous word আকর্কশ, whereas the actual correction is অকর্কশ. Once again, even if the correction is a valid word, it fails to generate the actual correct word due to imprecise mask prediction. The detector module predicted the mask as __কর্ক_ and the purificator module further purified it to __কর্কশ, where the actual mask should be অ_কর্কশ, consequently leading to an imprecise correction.

| **Homonym Error** | **Typographical Deletion** |
|---|---|
| Input: আহূতি | Input: উত্তারাধকারসূত্রে |
| (DTransformer) আহুতি ✓ | (DTransformer) উত্তারাধকারসূত্রে ✗ |
| (DCSpell) আহূত ✗ | (DCSpell) উত্তারাধকারসূত্রে ✗ |
| (DPCSpell) আহত ✗* | (DPCSpell) উত্তারাধিকারসূত্রে ✓ |
| **Run-on Error** | **Visual Error (Combined)** |
| Input: ক্ষতিপূরণার্থঅলাগ | Input: ঝাঁজবিশিস্ট |
| (DTransformer) ক্ষতিপূরণার্থ ✓ | (DTransformer) লাফজবিশিস্ট ✗ |
| (DCSpell) ক্ষতিপূরণার্থঅলাপ ✗ | (DCSpell) ঝাঁজবিশিষ্ট ✓ |
| (DPCSpell) ক্ষতিপূরণার্থ ✓ | (DPCSpell) ঝাঁজবিশিষ্ট ✓ |
| **Split-word Error (both)** | **Typographical Insertion** |
| Input: ত্রিরাশি | Input: সফললতার |
| (DTransformer) ত্রি রাশি ✓ | (DTransformer) সফলতার ✓ |
| (DCSpell) ত্রি রাশি ✓ | (DCSpell) সঁলতার ✗ |
| (DPCSpell) ত্রি রাশি ✓ | (DPCSpell) সফলতার ✓ |

Table 7: Comparison of our proposed DPCSpell with other competitive methods in relation to the qualitative result of Bangla SEC task

### 5.5.3 Spelling Error Correction for Resource-Scarce Indic Languages

We compare the performance of our method with three other tactics including LSTMSeq2Seq, DTransformer, and DCSpell in two low-resource Indic languages namely Hindi and Telugu, which are Sanskrit-oriented and follow left-to-right typing script like Bangla [2]. The performance of these methods can be found in table 8. Despite the fact that DTransformer performs incredibly well in both languages, our proposed DPCSpell achieves competitive performance with it. Most importantly, we notice that our method starts outperforming other approaches when it has a sufficient amount of training data. Initially, it was the worst performing method for Hindi and Telugu when training on a tiny corpus of Hindi and Telugu respectively from [7]. Afterwards, we enhance these corpora by incorporating nine types of spelling errors, utilizing our corpus creation tactic. Consequently, it outperforms LSTMSeq2Seq and DCSpell for both Hindi and Telugu SEC task. It suppresses the performance of DCSpell for Hindi SEC by EM score of 1.13%, and PR, RE, F1 and F0.5 scores of $8.5 \times 10^{-3}$, $1.13 \times 10^{-2}$, $1.39 \times 10^{-2}$, $6.2 \times 10^{-3}$. Meanwhile, it attains more competitive performance with DTransformer while outperforms DCSpell for Telugu SEC by improving its prior EM, PR, RE, F1, and F0.5 scores by 1.83%, $5.84 \times 10^{-2}$, $1.83 \times 10^{-2}$, $2.39 \times 10^{-2}$, $3.82 \times 10^{-2}$, respectively.

In comparison to our method's performance on Bangla where DPCSpell outperforms all the listed methods including DTransformer, it appears to suppress the performance of other methods on large-scale corpora as it shows a lucid improvement in its performance on the enhanced corpora. Since the enhanced Hindi and Telugu corpus is 6.54% and 5.64% times smaller than the Bangla corpus, it will perform even better in the Hindi and Telugu languages for large enough corpora.

| **Method** | **EM** | **PR** | **RE** | **F1** | **F0.5** | **Corpus** Lang. |
|---|---|---|---|---|---|---|
| LSTMSeq2Seq [7] | 85.40% | – | – | – | – | Hindi |
| **DTransformer [45]** | **90.43%** | **0.906** | **0.9043** | **0.9066** | **0.9075** | Hindi |
| DCSpell [38] | 82.18% | 0.8724 | 0.8218 | 0.8386 | 0.8562 | Hindi |
| DPCSpell | 78.64% | 0.8431 | 0.7864 | 0.8207 | 0.8238 | Hindi |
| **DTransformer [45]** | **96.71%** | **0.976** | **0.9671** | **0.9663** | **0.976** | Hindi* |
| DCSpell [38] | 85.80% | 0.9588 | 0.8580 | 0.8912 | 0.9248 | Hindi* |
| DPCSpell | 86.93% | 0.9673 | 0.8693 | 0.9051 | 0.9310 | Hindi* |
| LSTMSeq2Seq [7] | 89.30% | – | – | – | – | Telugu |
| **DTransformer [45]** | **95.66%** | **0.9587** | **0.9566** | **0.9585** | **0.9593** | Telugu |
| DCSpell [38] | 91.05% | 0.9225 | 0.9105 | 0.9203 | 0.9256 | Telugu |
| DPCSpell | 88.58% | 0.9058 | 0.8858 | 0.9008 | 0.9066 | Telugu |
| **DTransformer [45]** | **98.88%** | **.9872** | **.9888** | **.9899** | **.991** | Telugu* |
| DCSpell [38] | 89.91% | 0.9629 | 0.8991 | 0.9209 | 0.9422 | Telugu* |
| DPCSpell | 90.41% | 0.9642 | 0.9041 | 0.9247 | 0.9448 | Telugu* |

Table 8: The comparison of the quantitative outcomes of our proposed DPCSpell with other competitive methods for resource-scarce Indic languages such as Hindi and Telugu, where $*$ indicates the enhanced corpus.

## 5.6 Ablation Study

In this subsection, we investigate the impact of several DPCSpell components including the effect of the detector and purification module, masked characters, and beam search decoding for Bangla spelling error correction.

| **Method** | Mask | Correction | | | | | |
|---|---|---|---|---|---|---|---|
| | **EM** | **EM** | **PR** | **RE** | **F1** | **F0.5** | **MA** |
| $\mathcal{C}$ | – | 90.44% | 0.9061 | 0.9044 | 0.9047 | 0.9056 | 91.12% |
| $\mathcal{D}+\mathcal{C}$ | 88.54% | 84.23% | 0.8458 | 0.8423 | 0.8434 | 0.8446 | 85.07% |
| $\mathcal{D}+\mathcal{P}+\mathcal{C}$ | 96.86% | 94.78% | 0.9487 | 0.9478 | 0.948 | 0.9483 | 95.16% |

Table 9: Impact of different components of our proposed DPCSpell on the performance of the Bangla SEC task

### 5.6.1 Effect of the Detector Network

The detector module identifies the position of the erroneous characters in the input word and replaces them with a mask. As a result, the corrector module can only rectify the erroneous characters rather than the entire sequence. Furthermore, the corrector module becomes heavily reliant on the detector module's efficacy for the same reason. From table 9, we observe that the corrector module without the detector ($\mathcal{C}$), which is analogous to the DTransformer[45], performs better than the detector and corrector modules together ($\mathcal{D} + \mathcal{C}$), which is similar to the DCSpell[38], because the detector ($\mathcal{D}$) fails to precisely identify the erroneous characters of the word which consequently misleads the corrector ($\mathcal{C}$) and degrades its performance.

### 5.6.2 Effect of the Purificator Network

The purificator ($\mathcal{P}$) detects the erroneous letters in a word by further purifying the masked output of the detector ($\mathcal{D}$), which in turn allows the corrector ($\mathcal{C}$) to amend the erroneous portion only rather than the whole word. Table 9 reveals that the purificator improves the EM score of the detector module by 8.32%. It detects the masks with an EM score of 96.86% which indicates its effectiveness in mask detection. As a consequence, the corrector network's performance is significantly improved. The purificator's efficacy improves the corrector's EM score by 10.55% and helps to outperform previous methods by a convincing margin. It achieves an EM score of 94.78% in correction generation, whereas it was 84.23% and 90.44% in corrector with and without detector model, respectively.

### 5.6.3 Effect of Masked Characters

Figure 4 depicts how the model converges relatively faster due to precisely masking the erroneous characters. The loss of the Detector+Purificator+Corrector is significantly reduced between 5 and 15 epochs, whilst other variants exhibited gradual decrement. It signifies the method's aptitude to generate accurate correction in a short time frame. It takes approximately half of the time of the variant named corrector to reach the lowest value of the loss. Likewise, it hits the lowest point three times faster than the remaining detector+corrector variant.

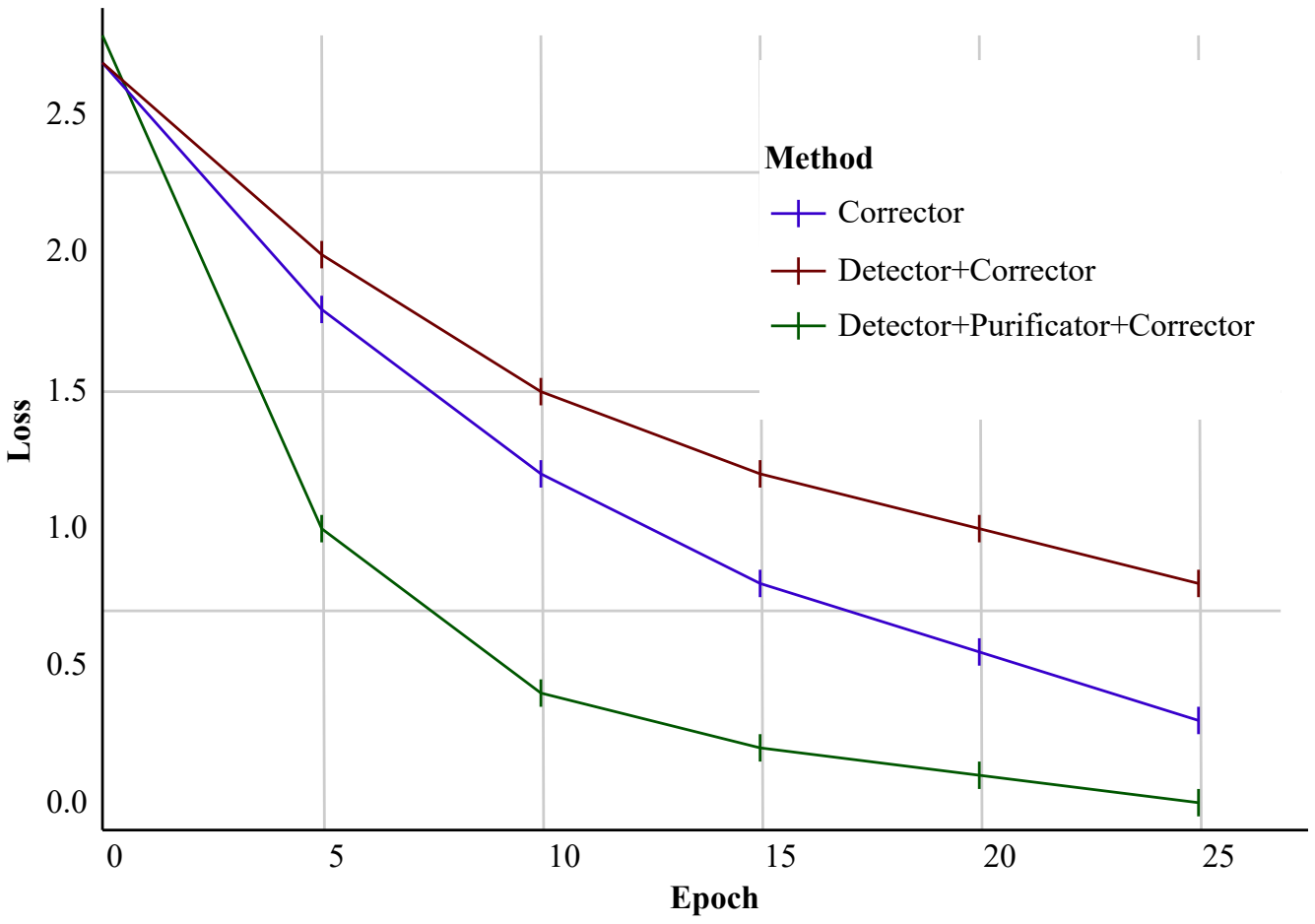

Figure 4: Effect of purified masked characters from the purificator network

### 5.7 Drawbacks of DPCSpell

Even though our proposed DPCSpell gives promising performance by outperforming several methods, it suffers from two minor issues. Firstly, it has a large parameter size as compared to other SEC methods. Each of the transformer-based detector, purificator, and corrector networks has 1,696,197 trainable parameters. As a result, it contains nearly 150% and 300% more trainable parameters than the DCSpell and DTransformer, respectively. Consequently, it requires almost twice and thrice as long as DCSpell[38] and DTransformer[45] to make a correction. Secondly, it is heavily data-dependent to produce a credible result. The experimental findings in table 8 delineate how our method begins to show its efficacy for a larger corpus. However, in the case of immense parameter size, the advance in technology helps overcome the hurdles in training a model with a parameter size of approximately 5M, which is not even end-to-end. Regarding the data dependencies, we propose a method for developing a large-scale corpus that is effective in resolving the data dependency issue.

## 6 Conclusion

The spelling error rectification task becomes challenging due to the visual and phonological features of characters which give ambiguous information about the context and essentially mislead the model. To solve the problem, we proposed a detector-purificator-corrector framework DPCSpell based on denoising transformers that detects whether a letter is appropriate or not before correcting it. The detector network is used to identify the erroneous characters and mask them, while the purificator further purifies the masked output. The corrector module is responsible for correction generation. We divided the SEC task into three sub-tasks, which significantly enhanced overall performance. Consequently, it became the new state-of-the-art method for Bangla SEC. In addition, we presented a new approach for creating a large-scale parallel corpus for SEC of any left-to-right scripted language which in turn resolved the resource limitation issue. A large-scale parallel corpus for Bangla SEC is developed using our method and made publicly available, making Bangla a resourceful language for the task. Furthermore, we observed that many existing methods rely on private corpora and withhold their codes, hindering reproducibility. So, we have made all our codes publicly available, fostering a reproducible baseline for the task. In the future, we will make our method less data-dependent with the help of meta-learning.

## Acknowledgement

This research is funded by Institute of Advanced Research (Grant No. UIU/IAR/02/2021/SE/22), United International University, Bangladesh.

## References

[1] Michael Flor, Yoko Futagi, Melissa Lopez, and Matthew Mulholland. Patterns of misspellings in l2 and l1 english: A view from the ets spelling corpus. *Bergen language and linguistics studies*, 6, 2015.

[2] Chowdhury Rafeed Rahman. Neural language modeling for context based word suggestion, sentence completion and spelling correction in Bangla. Department of computer Science and Engineering, Bangladesh University of Engineering and Technology, 2021. `http://lib.buet.ac.bd:8080/xmlui/bitstream/handle/123456789/5991/Full%20Thesis.pdf`.

[3] Muhammad Ifte Khairul Islam, Rahnuma Islam Meem, Faisal Bin Abul Kasem, Aniruddha Rakshit, and Md Tarek Habib. Bangla spell checking and correction using edit distance. In *2019 1st International Conference on Advances in Science, Engineering and Robotics Technology (ICASERT)*, pages 1–4. IEEE, 2019.

[4] Istiak Ahamed, Maliha Jahan, Zarin Tasnim, Tajbia Karim, SM Salim Reza, and Dilshad Ara Hossain. Spell corrector for bangla language using norvig's algorithm and jaro-winkler distance. *Bulletin of Electrical Engineering and Informatics*, 10(4):1997–2005, 2021.

[5] Prabhakar Gupta. A context-sensitive real-time spell checker with language adaptability. In *2020 IEEE 14th International Conference on Semantic Computing (ICSC)*, pages 116–122. IEEE, 2020.

[6] Antara Pal, Sourav Mallick, and Alok Ranjan Pal. Detection and automatic correction of bengali misspelled words using n-gram model. In *2021 International Conference on Advances in Electrical, Computing, Communication and Sustainable Technologies (ICAECT)*, pages 1–5. IEEE, 2021.

[7] Pravallika Etoori, Manoj Chinnakotla, and Radhika Mamidi. Automatic spelling correction for resource-scarce languages using deep learning. In *Proceedings of ACL 2018, Student Research Workshop*, pages 146–152, 2018.

[8] Naushad UzZaman and Mumit Khan. A comprehensive bangla spelling checker. BRAC University, 2006. `https://dspace.bracu.ac.bd/xmlui/bitstream/handle/10361/668/A%20comprehensive%20Bangla%20spelling%20checker.pdf`.

[9] Tanmoy Debnath, Sumaiya Sajnin, and Md Montaser Hamid. A hybrid approach to design automatic spelling corrector and converter for transliterated bangla words. In *2020 23rd International Conference on Computer and Information Technology (ICCIT)*, pages 1–6. IEEE, 2020.

[10] HM Mahmudul Hasan, Md Adnanul Islam, Md Toufique Hasan, Md Araf Hasan, Syeda Ibnat Rumman, and Md Najmus Shakib. A spell-checker integrated machine learning based solution for speech to text conversion. In *2020 third international conference on smart systems and inventive technology (ICSSIT)*, pages 1124–1130. IEEE, 2020.

[11] MD HAIBUR RAHMAN. *A SYSTEM FOR CHECKING SPELLING, SEARCHING NAME & PROVIDING SUGGESTIONS IN BANGLA WORD*. PhD thesis, United International University, 2018.

[12] Prianka Mandal and BM Mainul Hossain. Clustering-based bangla spell checker. In *2017 IEEE International Conference on Imaging, Vision & Pattern Recognition (icIVPR)*, pages 1–6. IEEE, 2017.

[13] Nahid Hossain, Salekul Islam, and Mohammad Nurul Huda. Development of bangla spell and grammar checkers: Resource creation and evaluation. *IEEE Access*, 9:141079–141097, 2021.

[14] Sourav Saha, Faria Tabassum, Kowshik Saha, and Marjana Akter. *BANGLA SPELL CHECKER AND SUGGESTION GENERATOR*. PhD thesis, United International University, 2019.

[15] Tanni Mittra, Sadia Nowrin, Linta Islam, and Deepak Chandra Roy. A bangla spell checking technique to facilitate error correction in text entry environment. In *2019 1st International Conference on Advances in Science, Engineering and Robotics Technology (ICASERT)*, pages 1–6. IEEE, 2019.

[16] Asmaul Hosna, Ayesha Khatun, Md Jahidul Islam, Md Mahin, Babe Sultana, and Sumaiya Kabir. Word clustering of bangla sentences using higher order n-gram language model. *GUB Journal of Science and Engineering (GUBJSE)*, 4(1):76–84, 2017.

[17] Md Mashod Rana, Mohammad Tipu Sultan, MF Mridha, Md Eyaseen Arafat Khan, Md Masud Ahmed, and Md Abdul Hamid. Detection and correction of real-word errors in bangla language. In *2018 International Conference on Bangla Speech and Language Processing (ICBSLP)*, pages 1–4. IEEE, 2018.

[18] Md Munshi Abdullah, Md Zahurul Islam, and Mumit Khan. Error-tolerant finite-state recognizer and string pattern similarity based spelling-checker for bangla. In *Proceeding of 5th international conference on natural language processing (ICON)*, 2007.

[19] Nur Hossain Khan, Gonesh Chandra Saha, Bappa Sarker, and Md Habibur Rahman. Checking the correctness of bangla words using n-gram. *International Journal of Computer Application*, 89(11), 2014.

[20] Tapashee Tabassum Urmi, Jasmine Jahan Jammy, and Sabir Ismail. A corpus based unsupervised bangla word stemming using n-gram language model. In *2016 5th International Conference on Informatics, Electronics and Vision (ICIEV)*, pages 824–828. IEEE, 2016.

[21] Amitava Das and Sivaji Bandyopadhyay. Morphological stemming cluster identification for bangla. *Knowledge Sharing Event-1: Task*, 3, 2010.

[22] Omar Sharif and Mohammed Moshiul Hoque. Automatic detection of suspicious bangla text using logistic regression. In *International Conference on Intelligent Computing & Optimization*, pages 581–590. Springer, 2019.

[23] Mir Noshin Jahan, Anik Sarker, Shubra Tanchangya, and Mohammad Abu Yousuf. Bangla real-word error detection and correction using bidirectional lstm and bigram hybrid model. In *Proceedings of International Conference on Trends in Computational and Cognitive Engineering*, pages 3–13. Springer, 2021.

[24] Soumik Sarker, Md Ekramul Islam, Jillur Rahman Saurav, and Md Mahadi Hasan Nahid. Word completion and sequence prediction in bangla language using trie and a hybrid approach of sequential lstm and n-gram. In *2020 2nd International Conference on Advanced Information and Communication Technology (ICAICT)*, pages 162–167. IEEE, 2020.

[25] Omor Faruk Rakib, Shahinur Akter, Md Azim Khan, Amit Kumar Das, and Khan Mohammad Habibullah. Bangla word prediction and sentence completion using gru: an extended version of rnn on n-gram language model. In *2019 International Conference on Sustainable Technologies for Industry 4.0 (STI)*, pages 1–6. IEEE, 2019.

[26] Sadidul Islam, Mst Farhana Sarkar, Towhid Hussain, Md Mehedi Hasan, Dewan Md Farid, and Swakkhar Shatabda. Bangla sentence correction using deep neural network based sequence to sequence learning. In *2018 21st International Conference of Computer and Information Technology (ICCIT)*, pages 1–6. IEEE, 2018.

[27] Shashank Singh and Shailendra Singh. Hindia: a deep-learning-based model for spell-checking of hindi language. *Neural Computing and Applications*, 33(8):3825–3840, 2021.

[28] Amita Jain, Minni Jain, Goonjan Jain, and Devendra K Tayal. "uttam" an efficient spelling correction system for hindi language based on supervised learning. *ACM Transactions on Asian and Low-Resource Language Information Processing (TALLIP)*, 18(1):1–26, 2018.

[29] Gurjit Kaur, Kamaldeep Kaur, and Parminder Singh. Spell checker for punjabi language using deep neural network. In *2019 5th International Conference on Advanced Computing & Communication Systems (ICACCS)*, pages 147–151. IEEE, 2019.

[30] Himadri Patel, Bankim Patel, and Kalpesh Lad. Jodani: A spell checking and suggesting tool for gujarati language. In *2021 11th International Conference on Cloud Computing, Data Science & Engineering (Confluence)*, pages 94–99. IEEE, 2021.

[31] Ahmad Ahmadzade and Saber Malekzadeh. Spell correction for azerbaijani language using deep neural networks. *arXiv preprint arXiv:2102.03218*, 2021.

[32] S Sooraj, K Manjusha, M Anand Kumar, and KP Soman. Deep learning based spell checker for malayalam language. *Journal of Intelligent & Fuzzy Systems*, 34(3):1427–1434, 2018.

[33] Romila Aziz and Muhammad Waqas Anwar. Urdu spell checker: A scarce resource language. In *International conference on intelligent technologies and applications*, pages 471–483. Springer, 2019.

[34] Borbála Siklósi, Attila Novák, and Gábor Prószéky. Context-aware correction of spelling errors in hungarian medical documents. *Computer Speech & Language*, 35:219–233, 2016.

[35] Charana Sonnadara, Surangika Ranathunga, and Sanath Jayasena. Sinhala spell correction a novel benchmark with neural spell correction.

[36] Upuli Liyanapathirana, Kaumini Gunasinghe, and Gihan Dias. Sinspell: A comprehensive spelling checker for sinhala. *arXiv preprint arXiv:2107.02983*, 2021.

[37] Mehedi Hasan Bijoy, Nirob Hasan, Md Tahrim Faroque Tushar, and Shafin Rahmany. Image tagging by fine-tuning class semantics using text data from web scraping. In *2021 24th International Conference on Computer and Information Technology (ICCIT)*, pages 1–6. IEEE, 2021.

[38] Jing Li, Gaosheng Wu, Dafei Yin, Haozhao Wang, and Yonggang Wang. Dcspell: A detector-corrector framework for chinese spelling error correction. In *Proceedings of the 44th International ACM SIGIR Conference on Research and Development in Information Retrieval*, pages 1870–1874, 2021.

[39] Ashish Vaswani, Noam Shazeer, Niki Parmar, Jakob Uszkoreit, Llion Jones, Aidan N Gomez, Łukasz Kaiser, and Illia Polosukhin. Attention is all you need. *Advances in neural information processing systems*, 30, 2017.

[40] Shaohua Zhang, Haoran Huang, Jicong Liu, and Hang Li. Spelling error correction with soft-masked bert. *arXiv preprint arXiv:2005.07421*, 2020.

[41] Noura Farra, Nadi Tomeh, Alla Rozovskaya, and Nizar Habash. Generalized character-level spelling error correction. In *Proceedings of the 52nd Annual Meeting of the Association for Computational Linguistics (Volume 2: Short Papers)*, pages 161–167, 2014.

[42] Jiaying Xie, Kai Dang, Jie Liu, and Enlei Liang. Abc-fusion: Adapter-based bert-level confusion set fusion approach for chinese spelling correction. *Computer Speech & Language*, page 101540, 2023.

[43] Dzmitry Bahdanau, Kyunghyun Cho, and Yoshua Bengio. Neural machine translation by jointly learning to align and translate. *arXiv preprint arXiv:1409.0473*, 2014.

[44] Jonas Gehring, Michael Auli, David Grangier, Denis Yarats, and Yann N Dauphin. Convolutional sequence to sequence learning. In *International conference on machine learning*, pages 1243–1252. PMLR, 2017.

[45] Alex Kuznetsov and Hector Urdiales. Spelling correction with denoising transformer. *arXiv preprint arXiv:2105.05977*, 2021.